\documentclass{llncs}
\usepackage{multirow}
\usepackage{bbding}
\usepackage[misc]{ifsym}
\usepackage{verbatim}
\usepackage{tabularx}
\usepackage{graphicx}
\usepackage{amsmath}
\usepackage{indentfirst}

\begin{document}
\title{3D Global Convolutional Adversarial Network\\ for Prostate MR Volume Segmentation}
\author{Haozhe Jia\inst{1,2} \and Yang Song\inst{2} \and Donghao Zhang\inst{2} \and Heng Huang\inst{3} \and Dagan Feng\inst{2} \and Michael Fulham\inst{4} \and Yong Xia\inst{1} \and Weidong Cai\inst{2}}
\institute{Shaanxi Key Lab of Speech \& Image Information Processing (SAIIP), School of Computer Science and Engineering, Northwestern Polytechnical University, Xi'an, PR China,710072 \and Biomedical and Multimedia Information Technology (BMIT) Research Group, School of Information Technologies, University of Sydney, NSW 2006, Australia \and Department of Electrical and Computer Engineering, University of Pittsburgh \and Department of PET and Nuclear Medicine, Royal Prince Alfred Hospital, NSW 2050, Australia}
\maketitle

\begin{abstract}
Advanced deep learning methods have been developed to conduct prostate MR volume segmentation in either a 2D or 3D fully convolutional manner. However, 2D methods tend to have limited segmentation performance, since large amounts of spatial information of prostate volumes are discarded during the slice-by-slice segmentation process; and 3D methods also have room for improvement, since they use isotropic kernels to perform 3D convolutions whereas most prostate MR volumes have anisotropic spatial resolution. Besides, the fully convolutional structural methods achieve good performance for localization issues but neglect the per-voxel classification for segmentation tasks. In this paper, we propose a 3D Global Convolutional Adversarial Network (3D GCA-Net) to address efficient prostate MR volume segmentation. We first design a 3D ResNet encoder to extract 3D features from prostate scans, and then develop the decoder, which is composed of a multi-scale 3D global convolutional block and a 3D boundary refinement block, to address the classification and localization issues simultaneously for volumetric segmentation. Additionally, we combine the encoder-decoder segmentation network with an adversarial network in the training phrase to enforce the contiguity of long-range spatial predictions. Throughout the proposed model, we use anisotropic convolutional processing for better feature learning on prostate MR scans. We evaluated our 3D GCA-Net model on two public prostate MR datasets and achieved state-of-the-art performances.
\end{abstract}

\section{Introduction}
\noindent Efficient detection and segmentation of the gland capsule in prostate MR images are critical for diagnosis, management, and prognosis. The traditional approaches on automatic prostate segmentation in MR images are mainly based on anatomical atlas registration \cite{klein2010elastix}, deformable models \cite{tsai2003shape}, and optimization algorithms \cite{qiu2014prostate}. However, the task still faces great challenges since the prostate MR images tend to come with large variability in the size / shape of the gland, heterogeneity in signal intensity around endorectal coils, and low contrast between the gland and adjacent structures.
\begin{figure}[!tbp]
\centering
\includegraphics[width=1\textwidth]{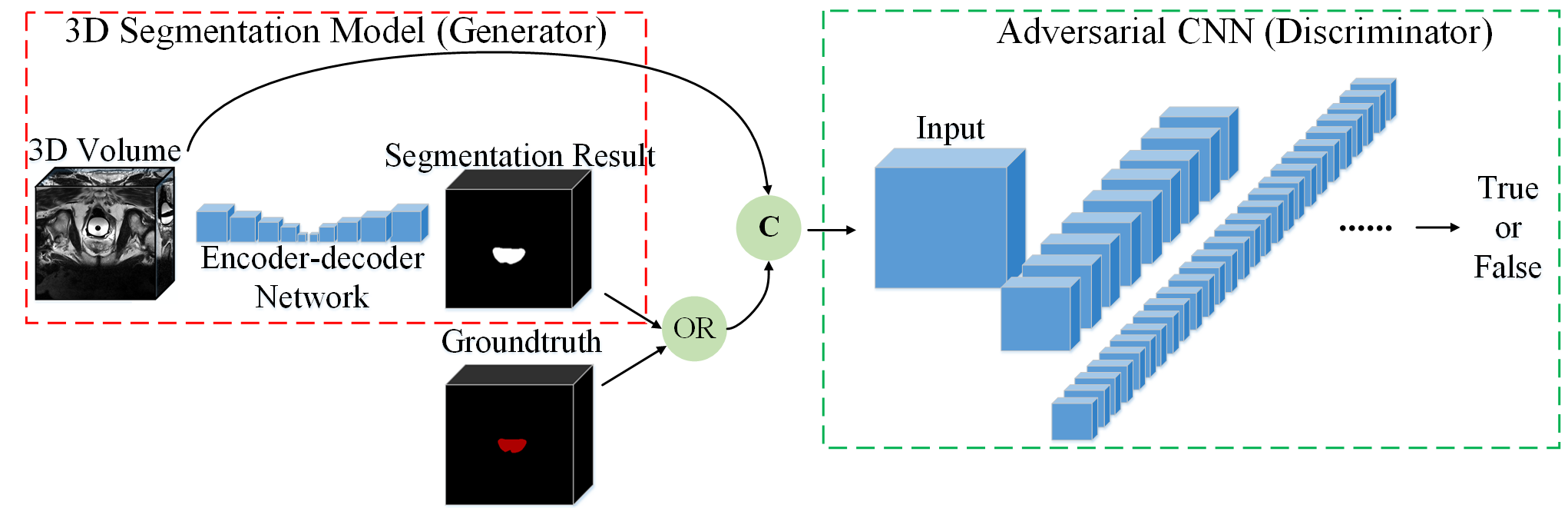}
\caption{Overview of the proposed approach}
\end{figure}
The availability of a large amount of annotated medical image data and pre-trained deep models has made it feasible to use deep learning for medical image segmentation and classification. Recently some deep convolutional neural network (CNN) based methods have achieved convincing segmentation performance in prostate MR images. A 2D approach \cite{drozdzal2018learning} was built based on combining the Fully Convolutional Network (FCN) \cite{long2015fully} and Residual Network (ResNet) \cite{he2016deep} for efficient prostate segmentation. Similar to the popular U-Net \cite{ronneberger2015unet}, a 3D segmentation model \cite{milletari2016vnet} was designed based on a volumetric, fully-convolutional neural network. Despite using volumetric convolutions and residual connections to maintain the spatial and contextual information, \cite{milletari2016vnet} ignores that the prostate MR volumes tend to come with anisotropic voxel resolution and especially have low between-slice resolution. In addition, since the biomedical volumetric segmentation task can be regarded as a dense per-voxel classification problem, both classification and localization issues are crucial for accurate segmentation. However, existing methods mainly focus on the localization issues but neglect the per-voxel classification problem, which tend to limit the segmentation performance.\\
\indent In this paper, we propose a new deep 3D Global Convolutional Adversarial Network (GCA-Net) that combines a 3D global convolutional network with an adversarial network for efficient prostate segmentation. Fig. 1 outlines the main components of our 3D GCA-Net. To address volumetric prediction of the prostate MR images, we first design a 3D fully convolutional encoder-decoder model for the segmentation task. We construct a 3D ResNet model \cite{he2016deep} as the encoder. In the decoder part, we design multi-level hybrid global convolution blocks and boundary refinement blocks to tackle classification and localization issues simultaneously for volumetric segmentation. In addition, to adaptive to the imaging resolution in prostate MR volumes, the traditional 2D convolution kernels are replaced by 3D anisotropic convolutions \cite{liu2017anisotropic} throughout the segmentation network. The multi-scale concatenation structure is also added to further reduce the loss of volumetric context information between different layers. Moreover, for CNN-based architectures, the spatial continuous prediction label maps are almost unavailable since the label variables are predicted independently by CNN from each other. To overcome this, in the training stage, we introduce adversarial training \cite{luc2016semantic} to the segmentation network as a regularization term. Compared to the traditional approaches \cite{alansary2016fast} which use conditional random fields (CRFs) to reinforce contiguity in the prediction label maps with pair-wise terms, the proposed 3D adversarial binary classification network can enforce the CNN to form higher-order consistent prediction maps, but also add no extra complexity to the segmentation model in the inference stage.\\
\indent We quantitatively and qualitatively evaluate our approach against several state-of-the-art approaches  \cite{milletari2016vnet,drozdzal2018learning,ronneberger2015unet} on both the MICCAI Grand Challenge-Prostate MR Image Segmentation Challenge 2012 (PROMISE 12) dataset \cite{litjens2014evaluation} and the NCI-ISBI 2013 Challenge-Automated Segmentation of Prostate Structures (ASPS 13) dataset \cite{challenge2013nci}. The experimental results show our 3D GCA-Net achieves superior segmentation performance compared to the state-of-the-art.
\begin{figure}[!tbp]
\centering
\includegraphics[width=1\textwidth]{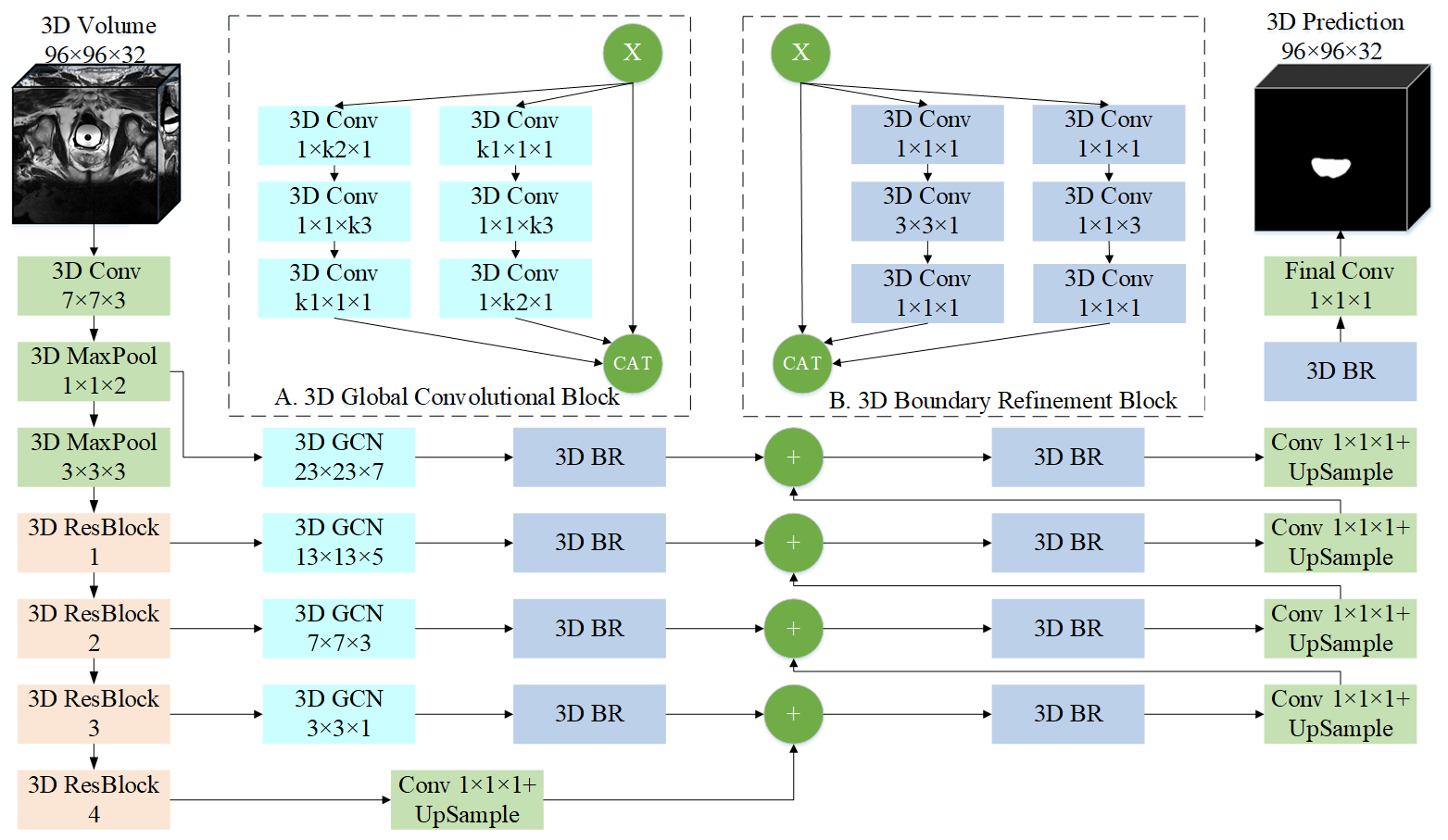}
\caption{\textbf{Architecture of 3D encoder-decoder segmentation network.} A and B are the 3D global convolutional block and 3D boundary refinement block of encoder, respectively. The sizes of input volume and the convolutional or pooling kernel of each layer are also shown.}
\end{figure}
\section{Method}
\subsection{3D Encoder-decoder Segmentation Network}
\noindent To achieve accurate and efficient segmentation of prostate MR volumes, we construct a 3D fully convolutional encoder-decoder deep network, in which the 3D ResNet encoder extracts the abundant context features of input volumes, and the 3D global convolutional and boundary refinement decoder further exploits the multi-scale features generated by the encoder to give the per-voxel prediction of the original input volumes.

\noindent\textbf{3D ResNet Encoder.}
We adopt the widely used ResNet-50 model \cite{he2016deep} as the base structure for our encoder, due to its superior ability to feature extraction. ResNet-50, however, is originally designed for 2D image processing. To extend it to 3D image segmentation, we replace its 2D convolutional layers with 3D ones. Specifically, in the ResNet-50 model, the first convolution layer with the kernel size of $7\times7$, is supposed to receive RGB 3 channel 2D input images. After extending ResNet-50 to our 3D model, the first convolutional layer has 3D kernels with the size of $7\times7\times3$, and therefore is able to receive 3D MR volumes. For all the other convolutional layers, we directly expand 2D convolutions to 3D convolutions with kernel size of 1 in z dimension. As a result, we extend ResNet-50 to a 3D ResNet encoder but retain compatibility with the original parameter setting. Hence it is still feasible for the 3D encoder to utilize the pre-trained parameters and transfer the knowledge about image representation learned on large scale natural images to characterize prostate MR images.\\
\noindent\textbf{3D Global Convolutional and Boundary Refinement Decoder.}
Inspired by the Global Convolutional Network (GCN) \cite{peng2017large} and aiming to achieve a sufficient exploitation of the 3D volumes features, we build the decoder with multi-scale 3D global convolution blocks and 3D boundary refinement blocks. Compared to conventional 3D convolutions with fixed and small kernel size, the 3D global convolution blocks use larger and multi-scale kernel convolutions to model the dense-connected structure of classification models, which can enhance the voxel classification capability of the network besides its original promising localization performance. Moreover, the 3D boundary refinement blocks use anisotropic and residual concatenation convolutions to further exploit both within-slice and between-slice features, which can achieve promising boundary localization of prostate MR volumes.\\
\indent In the 3D global convolutional block, we achieve large kernel size 3D convolution by decomposing it into a combination of three 1D convolutions on x-y-z dimensions, respectively, which performs dense connections on a large 3D block in the feature map but with limited
parameter numbers. As a result, we tackle the classification issues for the 3D volumetric semantic segmentation but also with limited computation cost. In addition, in consideration of the special anisotropic resolutions of prostate MR volumes, in our boundary refinement block, the input features are passed into 2 anisotropic convolutional layers \cite{liu2017anisotropic}, where the $3\times3\times1$ convolution further exploits the 2D features in x-y planes, and the $1\times1\times3$ convolution can focus on the features between different slices.
Additionally, in both the global convolutional block and the boundary refinement block, the residual concatenations are further added for a minimal loss of the feature information. The structures of the global convolutional block and the boundary refinement block are shown in Fig. 2A and 2B, respectively.\\
\indent In each stage of the decoder, the volumetric features extracted from the corresponding stage of the encoder will be passed to a 3D global convolutional block and a 3D boundary refinement block. Then, the output feature maps will be tri-linear upsampled and added with the higher scale ones. At the end of the decoder, we add a final $1\times1\times1$ convolution to generate the volumetric prediction as the segmentation of the input volume. The detailed components of the proposed segmentation network are shown in Fig. 2.
\subsection{Adversarial Training}
\noindent To further regularize the segmentation network to generate accurate and consistent volumetric prediction, we apply an adversarial training \cite{luc2016semantic} for our proposed segmentation network. In this approach, the adversarial learning is trained to detect the higher-order inconsistencies between ground truth and the segmentation result and guide the segmentation network to correct it, which is implemented with a hybrid loss function with the combination of a binary foreground-background weighted 3D cross-entropy segmentation loss term and an adversarial regularization term \cite{luc2016semantic}. We build a deep classification CNN with 6 convolutional layers but no fully connected layers as the discriminator model $D$. The objective functions of generator loss and discriminator loss are expressed as:
\begin{equation}
    \begin{split}
        Loss_G(X, Y)&=\lambda Loss_{Cross-Entropy}(G(X), Y)\\
        &+Loss_{GAN}(D(G(X)+X),True)\\
    \end{split}
\end{equation}
\begin{equation}
    \begin{split}
        Loss_D(G(X),Y)&=Loss_{GAN}(D(G(X)+X), False)\\
        &+Loss_{GAN}(D(Y+X), True)
    \end{split}
\end{equation}
where the generator $G$, i.e., the proposed segmentation network, is trained to generate segmentation result $G(X)$ that is similar to the ground truth $Y$ of the prostate MR volume $X$. In the meantime, $X$ is concatenated with $Y$ and $G(X)$, separately, then passed into the $D$ for a discrimination. It is noted that the adversarial training is only applied in the training stage and only the segmentation network is used to segment the volumes in the inference stage. Based on this, we can train the 3D segmentation model in a robust and effective manner but with no extra time consumption and computation complexity.
\section{Experiments and Results}
\noindent\textbf{Datasets:}
Two public prostate MR datasets were utilized to evaluate the proposed 3D GCA-Net. The first MICCAI PROMISE 12 challenge dataset \cite{litjens2014evaluation} contains 50 training transverse T2-weighted MR scans with corresponding annotated ground truth and 30 testing scans for online independent evaluation. The NCI-ISBI ASPS 13 challenge dataset \cite{challenge2013nci} consists of 60 MR scans, half of which are acquired with 1.5T machine and the other half with 3T machine. We trained the proposed method on the 50 training scans of PROMISE 12 dataset and submitted the segmentation results of the 30 testing data to the ongoing challenge. For the ASPS 13 challenge dataset, we randomly split all 60 training scans into 4 independent groups to conduct a 4-fold cross validation. Additionally, for an intuitive and quantitative evaluation, we also implemented a 3D version of U-Net \cite{ronneberger2015unet} and V-Net \cite{milletari2016vnet} with identical experiment setting for comparison. Three evaluation metrics were applied, dice similarity coefficient (DSC) \cite{litjens2014evaluation}, 95\% Hausdorff distance (95\%HD) \cite{litjens2014evaluation} and average boundary distance (ABD) \cite{litjens2014evaluation}.

\begin{table}[!tbp]
\footnotesize
\centering
\caption{Quantitative comparison with several variations of convolutional encoder-decoder networks on PROMISE 12 dataset. The evaluation results on all 3 metrics were obtained from the organizers. For DSC, higher values are better, for ABD and 95\% HD, lower values are better.}
\scalebox{0.85}{
\newcommand{\tabincell}[2]{\begin{tabular}{@{}#1@{}}#2\end{tabular}}
\begin{tabular}{l|c|c|c|c|c|c|c|c|c|c|c}
\hline
\multicolumn{1}{l|}{\multirow{2}*{Method}}	&\multicolumn{1}{c|}{\multirow{2}*{Type}}&\multicolumn{3}{c|}{DSC}  &\multicolumn{3}{c|}{ABD(mm)} &\multicolumn{3}{c|}{95\%HD(mm)} &\multicolumn{1}{c}{\multirow{2}*{Score}}\\
\cline{3-11}
\multicolumn{1}{c|}{}  &\multicolumn{1}{c|}{} &Whole &Base &Apex  &Whole &Base &Apex   &Whole &Base &Apex &\multicolumn{1}{c}{}\\
\hline
CAMP-TUM2 \cite{milletari2016vnet}   &3D   &0.869 &0.843 &0.844  &2.233 &2.458 &2.030	 &5.708  &5.835  &4.618    &82.39\\
\hline
UdeM 2D \cite{drozdzal2018learning}  &2D       &0.874 &0.849 &0.842  &2.171 &2.386 &2.070	 &6.124  &6.444  &4.705    &83.02\\
\hline
MBIOS       &2D    &0.881 &0.850 &0.847  &2.827 &2.204 &2.596   &10.543 &5.969  &6.494    &83.70\\
\hline
BDSLab      &3D    &0.883 &0.876 &0.798  &\textbf{1.864} &1.997&2.574  &5.341  &5.316  &6.312    &85.16\\
\hline
3D GCA-Net (ours)  &3D&0.889 &\textbf{0.877}&0.861  &1.901 &\textbf{1.969}&1.901   &\textbf{4.990} &\textbf{4.703} &4.300  &85.20\\
\hline
CREATIS     &2D/3D &0.893 &0.866 &\textbf{0.868} &1.926 &2.135 &\textbf{1.742} &5.594 &5.620 &\textbf{4.222} &85.74\\
\hline
CUMED \cite{yu2017volumetric}       &3D    &\textbf{0.894} &0.864 &0.860  &1.950 &2.127 &1.744  &5.537 &5.407 &4.292 &\textbf{86.64}\\
\hline
\end{tabular}
}
\end{table}

\noindent\textbf{Implementation Details:}
We implemented the proposed method based on the Pytorch framework on a Linux system with an Intel 3.6GHz$\times$8 CPU, 32G memory and a 11G Nvidia GeForce 1080 Ti GPU. In pre-processing, for both training and testing scans, we performed N4 bias field correction \cite{tustison2010n4itk}, unified the voxel resolution to a fixed size of $1\times1\times1.5mm$ and normalized the intensity into zero mean and unit variance. During the training of the 3D GCA Net, due to the limited number of the training scans, we applied a multiple online data augmentation including random flipping (both up-down or left-right in x-y planes), random rotation (one of $\pm$25, 90, 180 and 270 degree in x-y planes), random Gaussian noise ($\sigma$ from 0.3 to 0.7). The input of the model was a 3D volumes with the size $96\times96\times32$ and batch size 2, which was also randomly extracted in an online manner. The 3D segmentation network and the CNN discriminator model were trained together, each using the Adam optimizer with the initial learning rate of $1e^{-3}$, betas of (0.9, 0.999), weight decay of $1e^{-6}$, and $\lambda = 100$. The weights of the 3D ResNet encoder were initialized with those of the pre-trained ResNet-50 model, other parameters in the model were randomly initialized. In the inference phase, we extracted the sub-volumes with a fixed stride of $48\times48\times16$ in each testing scan and averaged the corresponding output of these sub-volumes predicted by the segmentation network to get the final segmentation result.\\
\noindent\textbf{Performance of Prostate Segmentation:}
Fig. 3 represents the qualitative segmentation results of the proposed 3D GCA-Net on PROMISE 12 dataset. We can observe that the segmented boundary is very close to the real one on both base part (Case04-Slice16) and central part of the prostate gland. Table 1 shows the quantitative segmentation results achieved by six methods. From Table 1, we can find that our 3D GCA-Net is superior to CAMP-TUM2, UdeM 2D, MBIOS and BDSLab, and competitive with CREATIS and CUMED \cite{yu2017volumetric}. Since a smaller 95\%HD value means less outliers in segmentation results, we can see that the introduction of adversarial learning can generate more consistent and smooth output label maps than other methods. We also find out that the methods implemented with 3D convolutions outperform those using 2D convolutions. This suggests that the volumetric convolutions should be applied to 3D prostate MR image segmentation. Furthermore, we compare our 3D GCA-Net and 3D encoder-decoder with two other widely used 3D fully convolutional segmentation networks by conducting 4-fold cross validation on the ASPS 13 dataset. The results shown in Table 2 demonstrate that our 3D encoder-decoder network equipped with global convolutional and anisotropic boundary refinement blocks achieves more accurate segmentation than V-Net and 3D U-Net, which use isotropic and small convolutional kernels. Besides, from Table 2 we can observe that our proposed 3D segmentation encoder-decoder has much more convolutional layers but less parameters compared to V-Net and 3D U-Net, which demonstrates the efficiency of our proposed method. Lastly, we further validate the effect of adversarial learning. Regarding the last two rows of Table 2, especially the ABD values, we can find that the adversarial network can further enforce the 3D encoder-decoder segmentation model to generate reasonable and consistent output label maps.
\begin{table}[!tp]
\footnotesize
\centering
\caption{Quantitative comparison of different methods on ASPS 13 dataset and the corresponding numbers of convolutional layers and parameters. ResNet-50 is shown here as a reference since our 3D ResNet encoder has the same numbers of convolutional layers and parameters with it.}
\scalebox{1}{
\begin{tabular}{l|c|c|c|c}
\hline
Method	     &Conv Layers     &Parameters       & DSC           &ABD(mm)\\ \hline
V-Net \cite{milletari2016vnet} &31  &65,191,134   &0.841         &2.531\\ \hline
3D U-Net \cite{ronneberger2015unet}   &23  &33,854,722  &0.862  &2.487\\ \hline
ResNet-50 \cite{he2016deep}  &53   &23,507,904  &/  &/ \\ \hline
3D Encoder-decoder (ours) &141  &29,601,094 &0.878    &2.402\\ \hline
3D GCA-Net (ours)  &148    &33,540,327   &\textbf{0.880} &\textbf{2.152} \\\hline
\end{tabular}}
\end{table}
\begin{figure}[!tp]
\centering
\includegraphics[width=0.9\textwidth]{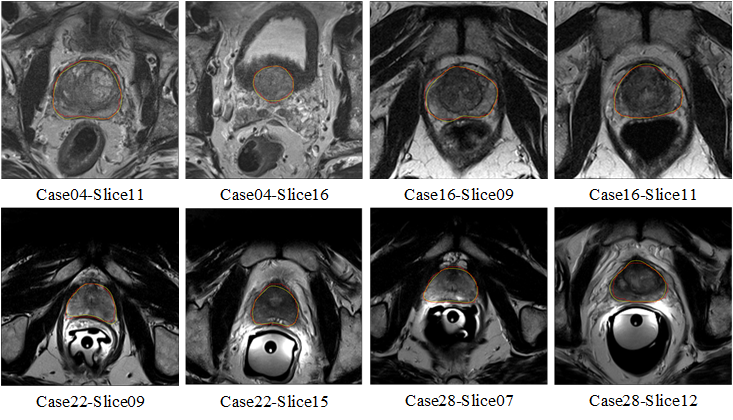}
\caption{Qualitative results of our proposed 3D GCA-Net on PROMISE 12 dataset, which were obtained from the organizers. The ground truth shown in yellow and segmentation result displayed in red.}
\end{figure}
\section{Conclusions}
\noindent In this paper, we propose a novel deep learning architecture called 3D GCA-Net for prostate MR volume segmentation. A 3D encoder-decoder segmentation network is first designed for the segmentation task, including a ResNet encoder for 3D prostate volume feature extraction and a multi-scale 3D global convolutional and boundary refinement decoder to successfully and simultaneously address both the classification and localization issues. Additionally, in the training phrase, an auxiliary adversarial network is introduced to the 3D segmentation network to further correct the segmentation results. The evaluation on two public MR prostate datasets demonstrates that our proposed approach improves the performance of the state-of-the-art.

\bibliographystyle{splncs03}
\bibliography{paper}

\begin{thebibliography}{10}
\providecommand{\url}[1]{\texttt{#1}}
\providecommand{\urlprefix}{URL }

\bibitem{challenge2013nci}
{NCI-ISBI} 2013 challenge. {A}utomated segmentation of prostate structures
  (2013)

\bibitem{alansary2016fast}
Alansary, A., Kamnitsas, K., Davidson, A., et~al.: Fast fully automatic
  segmentation of the human placenta from motion corrupted {MRI}. In: MICCAI.
  pp. 589--597. Springer (2016)

\bibitem{drozdzal2018learning}
Drozdzal, M., Chartrand, G., Vorontsov, E., et~al.: Learning normalized inputs
  for iterative estimation in medical image segmentation. MIA  44,  1--13
  (2018)

\bibitem{he2016deep}
He, K., Zhang, X., Ren, S., et~al.: Deep residual learning for image
  recognition. In: CVPR. pp. 770--778 (2016)

\bibitem{klein2010elastix}
Klein, S., Staring, M., Murphy, K., et~al.: Elastix: a toolbox for
  intensity-based medical image registration. IEEE TMI  29(1),  196--205 (2010)

\bibitem{litjens2014evaluation}
Litjens, G., Toth, R., van~de Ven, W., et~al.: Evaluation of prostate
  segmentation algorithms for {MRI}: the {PROMISE12} challenge. MIA  18(2),
  359--373 (2014)

\bibitem{liu2017anisotropic}
Liu, S., Xu, D., Zhou, S.K., et~al.: {3D} anisotropic hybrid network:
  Transferring convolutional features from {2D} images to {3D} anisotropic
  volumes. arXiv preprint arXiv:1711.08580  (2017)

\bibitem{long2015fully}
Long, J., Shelhamer, E., Darrell, T.: Fully convolutional networks for semantic
  segmentation. In: CVPR. pp. 3431--3440 (2015)

\bibitem{luc2016semantic}
Luc, P., Couprie, C., Chintala, S., et~al.: Semantic segmentation using
  adversarial networks. arXiv preprint arXiv:1611.08408  (2016)

\bibitem{milletari2016vnet}
Milletari, F., Navab, N., Ahmadi, S.A.: V-net: Fully convolutional neural
  networks for volumetric medical image segmentation. In: 3DV. pp. 565--571.
  IEEE (2016)

\bibitem{peng2017large}
Peng, C., Zhang, X., Yu, G., et~al.: Large kernel matters--improve semantic
  segmentation by global convolutional network. arXiv preprint arXiv:1703.02719
   (2017)

\bibitem{qiu2014prostate}
Qiu, W., Yuan, J., Ukwatta, E., et~al.: Prostate segmentation: an efficient
  convex optimization approach with axial symmetry using {3D TRUS} and {MR}
  images. IEEE TMI  33(4),  947--960 (2014)

\bibitem{ronneberger2015unet}
Ronneberger, O., Fischer, P., Brox, T.: U-net: Convolutional networks for
  biomedical image segmentation. In: MICCAI. pp. 234--241. Springer (2015)

\bibitem{tsai2003shape}
Tsai, A., Yezzi~Jr, A., Wells, W., et~al.: A shape-based approach to the
  segmentation of medical imagery using level sets. IEEE TMI  22(2),  137
  (2003)

\bibitem{tustison2010n4itk}
Tustison, N.J., Avants, B.B., Cook, P.A., et~al.: {N4ITK}: improved {N3} bias
  correction. IEEE TMI  29(6),  1310--1320 (2010)

\bibitem{yu2017volumetric}
Yu, L., Yang, X., Chen, H., Qin, J., Heng, P.A.: Volumetric convnets with mixed
  residual connections for automated prostate segmentation from 3d mr images.
  In: AAAI. pp. 66--72 (2017)

\end{thebibliography}
\end{document}